\documentclass{bmvc2k}

\title{One-shot Network Pruning at Initialization with Discriminative Image Patches}

\addauthor{Yinan Yang}{gr0528xf@ed.ritsumei.ac.jp}{1}
\addauthor{Yu Wang}{yu.wang@r.hit-u.ac.jp}{2}
\addauthor{Ying Ji}{jiying@nagoya-u.jp}{3}
\addauthor{Heng Qi}{hengqi@dlut.edu.cn}{4}
\addauthor{Jien Kato}{jien@fc.ritsumei.ac.jp}{1}

\addinstitution{
 Ritsumeikan University\\
 Shiga, Japan
}
\addinstitution{
	Hitotsubashi University\\
	Tokyo, Japan
}
\addinstitution{
 Nagoya University\\
 Nagoya, Japan
}
\addinstitution{
	Dalian University of Technology \\
	Dalian, China
}

\runninghead{YANG ET AL.}{OPaI with Discriminative Image Patches}

\def\etal{\emph{et al}\bmvaOneDot}

\usepackage{amssymb}
\usepackage{tikz}
\usepackage{pgfplots}
\usepackage[noend]{algpseudocode}
\usepackage{bbm}
\usepackage{algorithmicx,algorithm}
\usepackage{enumitem}
\usepackage{tabu}
\usepackage{graphicx,adjustbox}
\usepackage{verbatim}
\usepackage{multirow}
\usetikzlibrary{spy,backgrounds}
\def\x{{\mathbf x}}

\definecolor{butter1}{rgb}{0.988,0.914,0.310}
\definecolor{chocolate1}{rgb}{0.914,0.725,0.431}
\definecolor{chameleon1}{rgb}{0.541,0.886,0.204}
\definecolor{skyblue1}{rgb}{0.447,0.624,0.812}
\definecolor{plum1}{rgb}{0.678,0.498,0.659}
\definecolor{scarletred1}{rgb}{0.937,0.161,0.161}

\begin{document}

\maketitle

\begin{abstract}    
	One-shot Network Pruning at Initialization (OPaI) is an effective method to decrease network pruning costs. Recently, there is a growing belief that data is unnecessary in OPaI, e.g.\cite{Su_Chen_Cai_Wu_Gao_Wang_Lee_2020,Frankle_Dziugaite_Roy_Carbin_2020}. However, we obtain an opposite conclusion by ablation experiments in two representative OPaI methods, SNIP \cite{Lee_Ajanthan_Torr_2018} and GraSP \cite{Wang_Zhang_Grosse_2020}. Specifically, we find that informative data is crucial to enhancing pruning performance. In this paper, we propose two novel methods,  \textit{Discriminative One-shot Network Pruning (DOP)} and \textit{Super Stitching}, to prune the network by high-level visual discriminative image patches. Our contributions are as follows. (1) Extensive experiments reveal that OPaI is data-dependent. (2) Super Stitching performs significantly better than the original OPaI method on benchmark ImageNet, especially in a highly compressed model.
\end{abstract}

%-------------------------------------------------------------------------

\section{Introduction}

Since the 1980s, we have known that it is possible to substantially reduce parameters in neural networks without seriously compromising performance \cite{Reed_1993,Han_Pool_Tran_Dally_2015}. Such pruned neural networks can significantly decrease the computational demands of inference by using specific methods \cite{Gale_Zaharia_Young_Elsen_2020,Choquette_Gandhi_Giroux_Stam_Krashinsky_2020}. Among the various pruning methods developed so far, Network Pruning at Initialization (PaI) has attracted considerable attention since it provides a possibility to train sparse networks at lower costs \cite{Wang_Qin_Bai_Zhang_Fu_2021}. Specifically, PaI aims to achieve (close to) full accuracy (the accuracy reached by the dense network) by training a sparse subnetwork from a randomly initialized dense network. 

\begin{figure}[htb]
	\centering
	
	\centerline{\includegraphics[width=5in]{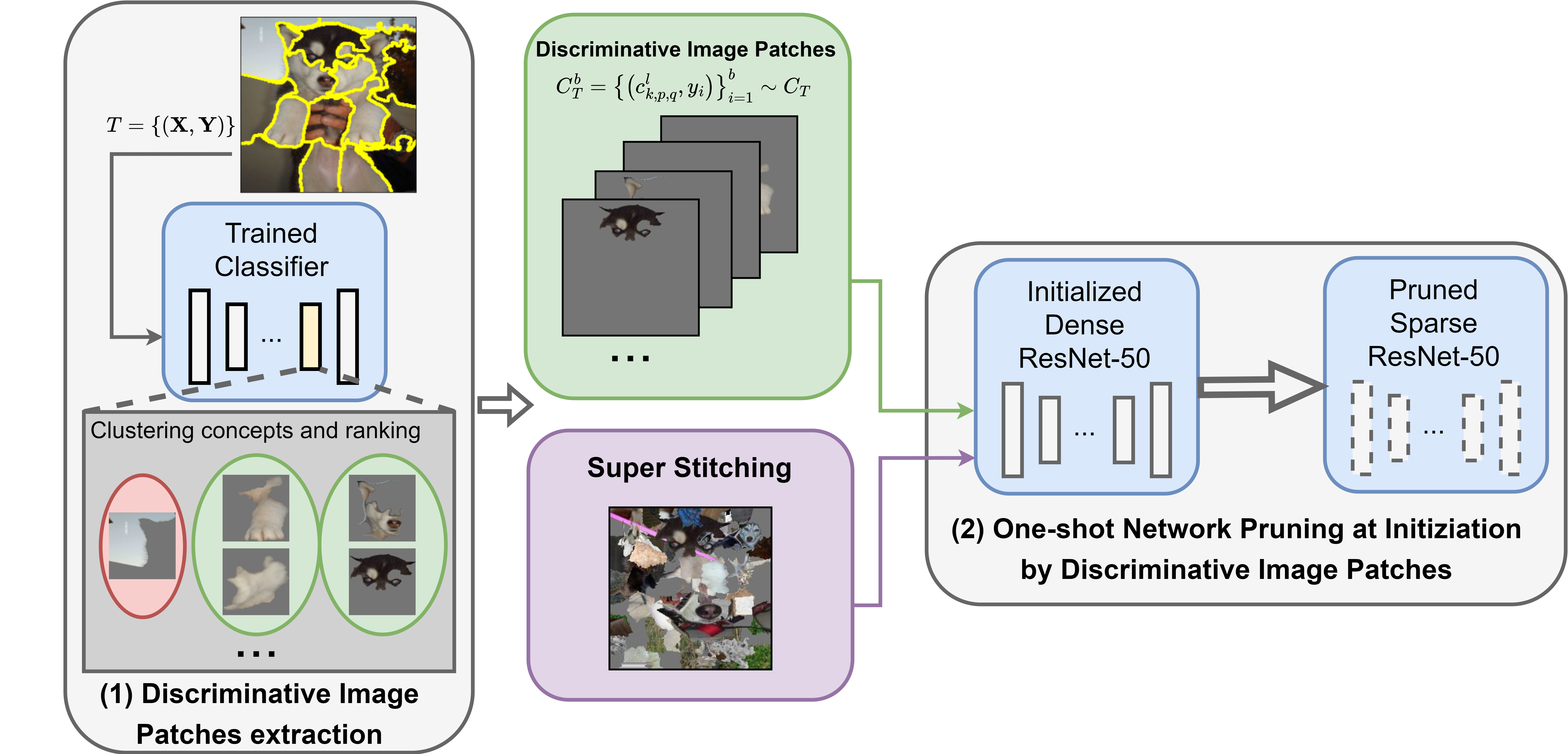}}
	\caption{\textbf{Overview of the Discriminative One-shot Network Pruning (DOP) and Super Stitching.} (1) Cluster segments in trained network's activation space, extract Discriminative Image Patches. The green is meaningful in network prediction, and the red is meaningless. (2) Using Discriminative Image Patches or Super Stitching to prune unimportant parameters by an specific OPaI algorithm.}
	\label{fig:Overview}
	%\vspace{-1em}
\end{figure}

In PaI research, the pruning criterion is the key focus \cite{Mozer_Smolensky_1989,Frankle_Carbin_2018,Molchanov_Tyree_Karras_Aila_Kautz_2016,LeCun_Denker_Solla_1989,Wang_Qin_Bai_Zhang_Fu_2021,Tanaka_Kunin_Yamins_Ganguli_2020,Mussay_Osadchy_Braverman_Zhou_Feldman_2019}. Most PaI works involve iterative pruning processing to improve performance while drastically increasing training costs.  In contrast, One-shot Network Pruning at Initialization (OPaI), another branch of PaI, attempts to reduce costs by single-step pruning. Specifically, SNIP \cite{Lee_Ajanthan_Torr_2018} and GraSP \cite{Wang_Zhang_Grosse_2020}, two representative methods of OPaI, use gradient information in the initial network to find subnetworks. Both algorithms employ random mini-batches in the pruning step, however, the data’s role has not been elucidated. Furthermore, despite the lack of extensive experimental evidence, there is a growing belief that data is not essential in OPaI \cite{Su_Chen_Cai_Wu_Gao_Wang_Lee_2020, Frankle_Dziugaite_Roy_Carbin_2020,Hoefler_Alistarh_Ben-Nun_Dryden_Peste_2021}, which may impact future OPaI or even PaI research.

%Under this mainstream, the importance of data preprocessing has been barely explored. 

This work questions the presumption of data independence in OPaI. To find the answer, we present Discriminative One-shot Network Pruning (DOP), as shown in Fig. \ref{fig:Overview}. Compared to previous studies, we employ discriminative data rather than random mini-batches. As a result, more precise gradient information is retained, so crucial structures and parameters are preserved in the network. To seek critical data in a trained classifier for targeted pruning, we distinctively generate discriminative image patches by Automatic Concept-based Explanation (ACE) \cite{Ghorbani_Wexler_Zou_Kim_2019}. In the ACE algorithm, visual concepts are vital for predicting a certain class and are automatically extracted for each class. The extracted visual concepts are processed into our discriminative image patches for subsequent pruning. Our extensive experiments on the benchmark ImageNet confirmed that discriminative data is signifiant to OPaI. Moreover, we propose an advanced strategy, Super Stitching, to further enhance the pruning performance. Super Stitching combines dozens of concepts in the same class but with different semantics. Such stitching patches are like chunks of condensed class information that enable pruning algorithms to get competitive outcomes with fewer samples. The experiment results on benchmark ImageNet show that Super Stitching is superior to the original SOTA OPaI method. Our major contributions are:

\begin{enumerate}
	%\vspace{-1em}
	\item Our experiments show that using discriminative data enhances pruning performance dramatically. We experimentally demonstrate that OPaI is data-dependent, which refreshes the knowledge of OPaI \cite{Su_Chen_Cai_Wu_Gao_Wang_Lee_2020,Frankle_Dziugaite_Roy_Carbin_2020}.
	% It indicates that the previous work of  \cite{Su_Chen_Cai_Wu_Gao_Wang_Lee_2020} still has non-negligible limitations.
	\item  We propose two novel data-dependent OPaI methods, DOP and Super Stitching. Super Stitching significantly improves the pruned network performance, especially in a highly compressed model. We use the One-Shot method with fewer samples to achieve higher or similar results than iterative pruning methods and SOTA. Our research demonstrates that, using informative data, One-Shot PaI can even outperform iterative pruning.
%	\vspace{-1em}
\end{enumerate}

%This paper proceeds as follows. Section \ref{relatedwork} introduces the related works. Section \ref{approach} reviews OPAI algorithms and introduces our proposed methods: DOP and Super Stitching. Section \ref{experiments} introduces our experiments on benchmark ImageNet, especially \ref{ssec:Ablations} demonstrates the data-dependent in OPAI. 

%-------------------------------------------------------------------------
\section{Related work}
\label{relatedwork}
\begin{comment}
	Our paper relates to two closed areas of research: (a) One-cut Network Pruning at Initialization, and (b) Data-dependent in OPAI. We discuss closely related work for each.
	\subsection{One-cut Network Pruning at Initialization (OPAI)}
	\subsection{Data-dependent in OPAI}
\end{comment}

Neural network pruning research started in the late 1980s \cite{Reed_1993}, and it is become increasingly popular until recently \cite{Han_Pool_Tran_Dally_2015,Wang_Qin_Bai_Zhang_Fu_2021}. There are three steps in a typical pruning pipeline: (1) pre-training, (2) pruning, and (3) fine-tuning \cite{Reed_1993,Han_Pool_Tran_Dally_2015}. Pruning has been regarded as a post-processing approach to solving the problem of redundancy \cite{Gale_Elsen_Hooker_2019,Blalock_Ortiz_Frankle_Guttag_2020,Hoefler_Alistarh_Ben-Nun_Dryden_Peste_2021,Liu_Sun_Zhou_Huang_Darrell_2018,Sze_Chen_Yang_Emer_2017,Cheng_Wang_Li_Hu_Lu_2018,Cheng_Wang_Zhou_Zhang_2018}. However, a novel pruning paradigm, Network Pruning at Initialization (PaI), is out of this pipeline \cite{Wang_Qin_Bai_Zhang_Fu_2021}, which trains a sparse subnetwork from a randomly initialized dense network instead of pruning a pre-trained network. PaI seeks to achieve full accuracy at lower training costs. One of the notable PaI efforts is the lottery ticket hypothesis (LTH) \cite{Frankle_Carbin_2018}, which selects subnetwork masks from pre-trained dense networks. This outstanding work sparks an interest in the deep learning community, and numerous related PaI researches have emerged \cite{Morcos_Yu_Paganini_Tian_2019,Malach_Yehudai_Shalev-Shwartz_Shamir_2020,Zhou_Lan_Liu_Yosinski_2019,Chen_Frankle_Chang_Liu_Zhang_Carbin_Wang_2020,You_Li_Xu_Fu_Wang_Chen_Baraniuk_Wang_Lin_2019, Wang_Qin_Bai_Zhang_Fu_2021}. 

%{\color{red}And according to a recent survey, the majority of focus in this line of PaI research lies in the pruning criterion} \cite{Wang_Qin_Bai_Zhang_Fu_2021}.

Most traditional PaI methods, such as LTH, dramatically raise the pruning cost since iterative pruning is needed. A distinctive branch of PaI, One-shot Network Pruning at Initialization (OPaI), solves this problem by computing the importance of each parameter in a neural network in a single step \cite{Wang_Qin_Bai_Zhang_Fu_2021,Wimmer_Mehnert_Condurache_2022,Hayou_Ton_Doucet_Teh_2020,Wimmer_Mehnert_Condurache_2021,Lubana_Dick_2020,Zhang_Stadie_2019,Alizadeh_Tailor_Zintgraf_Amersfoort_Farquhar_Lane_Gal_2022,subramaniam2022n2nskip,chen2021only}. Two basic OPaI methods are SNIP \cite{Lee_Ajanthan_Torr_2018} and GraSP \cite{Wang_Zhang_Grosse_2020}. Concretely, OPaI uses random mini-batches to calculate the connection sensitivity (SNIP) or the gradient signal preservation (GraSP) with respect to the loss for each parameter as an important score. Then, unimportant parameters are masked by a constant value of 0, and a trained sparse network is obtained by regular fine-tuning. 

%Furthermore, subsequent researches investigate the potential and limitations of OPAI \cite{Su_Chen_Cai_Wu_Gao_Wang_Lee_2020,Frankle_Dziugaite_Roy_Carbin_2020,Jorge_Sanyal_Behl_Torr_Rogez_Dokania_2020,Lee_Ajanthan_Gould_Torr_2019}.

A crucial concern has been raised about whether OPaI is data-dependent. For a long time, although it has been assumed that pruning methods use information from training data to find subnetworks, data in pruning has not received sufficient attention. Specially, Su \etal \cite{Su_Chen_Cai_Wu_Gao_Wang_Lee_2020} claim that data is unessential in SNIP and GraSP since the subnetworks generated by the corrupted data (dataset with random labels or pixels) behave similarly to the original one. However, such a perspective is not convincing because their pruning step actually still relies on information from the training set. More explicitly, their experiments cannot be considered fully data-independent, because the corrupted data retain the same distribution as the original training set. Recently, it is increasingly accepted that OPaI is independent of data, yet more experiments to support this conclusion are lacking. For instance, \cite{Frankle_Dziugaite_Roy_Carbin_2020} and \cite{Hoefler_Alistarh_Ben-Nun_Dryden_Peste_2021} discuss data independence as a limitation of OPaI. \cite{Vysogorets_Kempe_2021} develops Layerwise Sparsity Quotas based on the assumption of  data-independence. Nevertheless, in numerous benchmark tests, the different samples are one of the reasons for the disparities in outcomes \cite{Frankle_Dziugaite_Roy_Carbin_2020,Wang_Zhang_Grosse_2020,Jorge_Sanyal_Behl_Torr_Rogez_Dokania_2020}, which indicates that data matters in OPaI. To recap, the function of data in OPaI is still unclear and needs to be researched urgently.

%In addition, some researches claim that the pruning methods without any data performs better.

%In addition, some pruning methods \cite{Lee_Ajanthan_Gould_Torr_2019,Tanaka_Kunin_Yamins_Ganguli_2020,Frankle_Dziugaite_Roy_Carbin_2020} that do not require data reinforce the inherent notion that data is not currently essential at OPAI, even PAI. 

In contrast to earlier work that focused on pruning criteria, we investigate whether OPaI is data-dependent. Our work experimentally demonstrates that the improvement of OPaI performance requires informative data. From this viewpoint, we then propose our method, Super Stitching, and show it significantly outperforms the original OPaI approach on benchmark ImageNet. Our method beats the SOTA on ImageNet with a 95\% sparsity ResNet-50 model.

\section{Proposed Approaches}
\label{approach}
\label{sec:Proposed Method}

\textbf{Overview:} This section introduces our two approaches. Section \ref{ssec:revisiting} simply recalls two OPaI methods, SNIP and GraSP. They  prune network by random mini-batches at Initialization. Section \ref{COP_section} introduces our approach, DOP, which uses Discriminative Image Patches to replace random mini-batches in OPaI.  To improve the gradient information quality in pruning, we introduce Super Stitching  in Section \ref{Super_Stitching}.

\textbf{Notions:} We use $T=\{(\mathbf{X}, \mathbf{Y})\}$  to denote a certain classification task, where $x_{i} \in \mathbf{X}$ represents a sample, and $y_{i} \in  \mathbf{Y}$ represents its label. We consider a neural network $f\left ( T,\theta \right )$ before pruning for task $T$ with parameters $\theta$, and $\theta_{j}$ is the parameter of connection $j$ in $f\left ( T,\theta \right )$. An OPaI pruning criterion $\mathcal{A}$ produces $j$'s binary mask, denoted by $m_{j} \in \left\{0,1\right\}$. Given a sparsity level $\kappa$, the training of $f\left ( T,\theta \odot m \right )$ following the Minimization Empirical Risk:
\begin{equation}
	%\footnotesize
	\begin{gathered}
		\underset{m,\theta}{\arg \min } \frac{1}{n} \sum_{i}^{n} \mathcal{L}\left(\left(x_{i},f\left(\theta  \odot m \right)\right), y_{i}\right),\quad\|m\|_{0} \leq \kappa,
	\end{gathered}
\end{equation}
where $\mathcal{L}$ denotes the loss function, and $\odot$ denotes the Hadamard product.

\subsection{Recall SNIP and GraSP}
\label{ssec:revisiting}
In SNIP and GraSP, the first step is to sample mini-batches from the training set. The random mini-batch $T^{b}$ is 
\begin{equation}
	%\footnotesize
	\begin{gathered}
		T^{b}=\left\{\left(x_{i}, y_{i}\right)\right\}_{i=1}^{b} \sim T,
	\end{gathered}
	\label{g}
\end{equation}
which is used for computing gradient information. The important score $s(\theta_{j})$ of parameter $\theta_j$ in the two methods are as follows. 

\textbf{\textit{SNIP}} \cite{Lee_Ajanthan_Torr_2018} computes the connection sensitivity of each parameter $\theta_{j}$ as an important score. We use the right-hand formula \cite{Su_Chen_Cai_Wu_Gao_Wang_Lee_2020,Jorge_Sanyal_Behl_Torr_Rogez_Dokania_2020,Frankle_Dziugaite_Roy_Carbin_2020} to simplify the calculations:
\begin{equation}
	%\footnotesize
	\begin{gathered}
		s(\theta_{j})=\left|\frac{\partial L(T^{b};\theta_{j} \odot m_{j} )}{\partial m}\Big|_{m=1}\right|=\left|\frac{\partial L(T^{b}; \theta_{j})}{\partial \theta_{j}} \odot \theta_{j}\right|.
	\end{gathered}
	\label{g}
\end{equation}
%Once the sensitivity $|s(\theta_{j})|$ is computed, only the top-$\kappa$ connections are kept.
%The absolute value of the importance score $|s(\theta_{j})|$ serves as the basis for the retention of the parameter.

\textbf{\textit{GraSP}} \cite{Wang_Zhang_Grosse_2020} uses the Hessian $H$ to preserve the gradient flow in the final sparse network. The score is designed as follows:
\begin{equation}
	s(\theta_{j})=-\theta_{j}(H \frac{\partial L(T^{b}; \theta_{j})}{\partial \theta_{j}})_{j}.
\end{equation}

%{\color{red}Similar to the subsequent SNIP operation, GraSP prunes less important parameters and finally fine-tunes the pruned network with the mask.}

Once the importance scores of SNIP and GraSP are obtained, then, only the top-$\kappa$ connections are kept. And we finally fine-tunes the pruned network with the mask.

\subsection{Discriminative One-shot Network Pruning}
\label{COP_section}

\begin{figure*}[htbp]
	
	\centering
	\includegraphics[trim={0 0.1cm 0 0.2cm},clip,width=12.8cm]{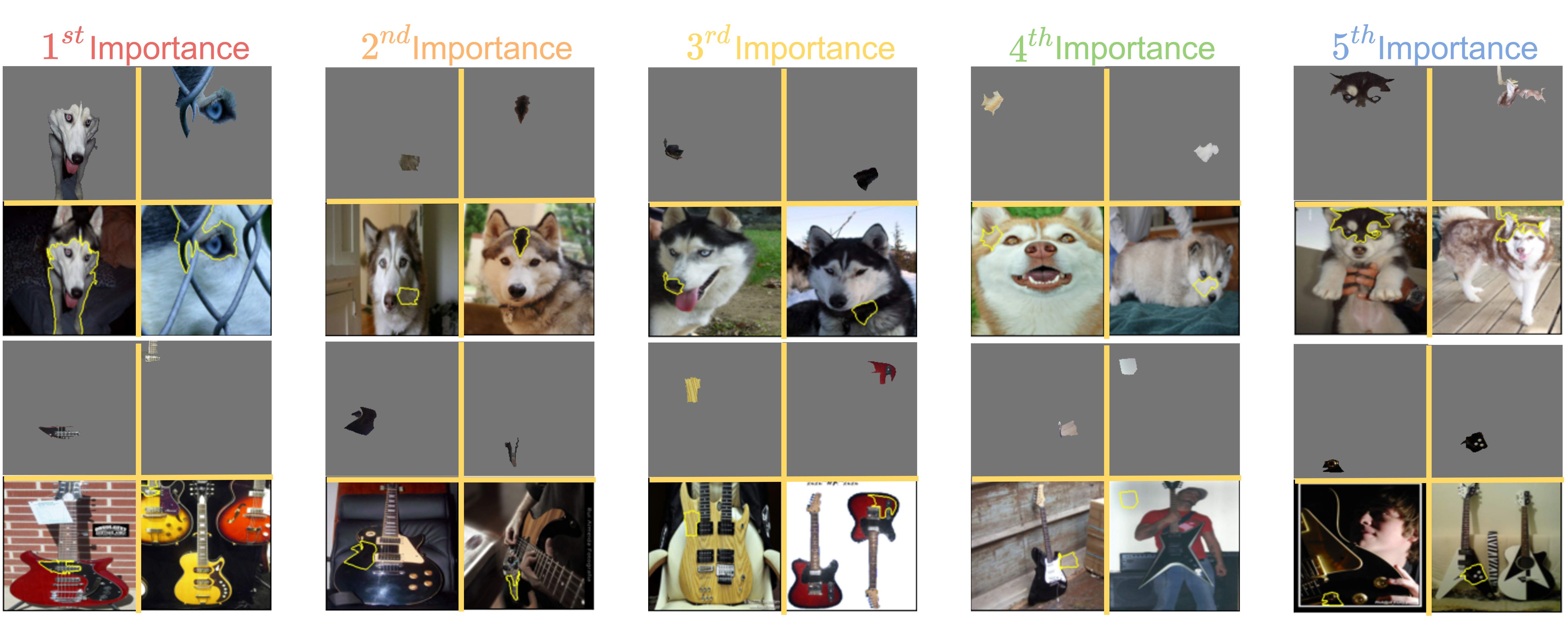}
	
	\caption{Discriminative image patches extracted by pre-trained ResNet-50 from the ImageNet. We choose the penultimate layer of ResNet-50 for the activation space. Here we show Siberian husky and electric guitar in each class's Top-5 important concepts based on their TCAV scores. The original images from the training set are shown below, and the discriminative image patches are shown above.}
	\label{concepts}
	%\vspace{-1.5em}
\end{figure*}
To replace $T^{b}$ with ``better data'' for pruning, we adopt ACE algorithm \cite{Ghorbani_Wexler_Zou_Kim_2019} to extract concepts from training data. Then these visual concepts are processed into our discriminative image patches. ACE adopts super-pixel segmentation on the training set,  clusters the activations in Euclidean Space,  and ranks the concepts by TCAV scores \cite{Kim_Wattenberg_Gilmer_Cai_Wexler_Viegas_Sayres_2017}. The higher the TCAV score, the more discriminative the concepts for the recognition of a certain class. Specifically, ACE needs a trained classifier $F( T )$ with a bottleneck layer $l$, as shown in Figure \ref{fig:Overview}. We denote $k$ as a class label in $T$, and $X_{k}$ is all inputs with that given label. We first segment images $X_{k}$ by a super-pixel algorithm called SLIC \cite{Achanta_Shaji_Smith_Lucchi_Fua_Su}. Then, we let the segments flow to $F( T )$, pick a bottleneck layer $l$ in $F( T )$ as activation space, and cluster similar segments in $l$ by K-Means. At that point, each harvested cluster plays a specific role in a target class, and we obtain a total of $p$ clusters, with $q$ independent segments in each cluster. Finally, we view such a cluster with the same semantics as a visual concept patch and unite a segment with a mean image value matrix (with the same size of $x_{i}$) to generate an image so-called discriminative image patch. We denote a set of discriminative image patches as $C_{T}\sim \{  (c_{k,p,q},y_{i} )| c_{k,p,q}\in x_{i} \}$, here $c_{k,p,q}$ means a discriminative image patch.
\begin{algorithm}[H]
	\caption{DOP: Discriminative One-shot Network Pruning}
	\begin{algorithmic}[1]
		\Require{Network $f\left ( T,\theta \right )$, loss function $\mathcal{L}$, sparsity level $\kappa$, a trained classifier $F\left ( T \right )$ with a bottleneck layer $l$, and an OPaI pruning criterion $\mathcal{A}$}
		\State $C_{T}\sim \left \{ \left ( c_{k,p,q},y_{i} \right ) \right \}\leftarrow \mathrm{ACE}\left ( F\left ( T \right ),l\right)$
		\Comment {Create Discriminative Image Patches}
		
		\State $C_{T}^{b}=\left\{\left ( c_{k,p,q},y_{i} \right )\right\}_{i=1}^{b} \sim C_{T}$
		\Comment {Sample mini-batches}
		\State $S\left ( \theta  \right )=\mathcal{A}\left ( C_{T}^{b},f\left ( T,\theta \right ) \right)$
		\Comment{pruning by $\mathcal{A}$ with Discriminative Image Patches}
		\State $m \leftarrow \mathbbm{1}\left[s_{\theta}-\tilde{s}_{\kappa} \geq 0\right]$
		\Comment{Make mask by keeping Top-$\kappa$ score}
		\State $f\left ( T,\theta \odot m  \right ) \leftarrow \arg \min _{\theta} \mathcal{L}\left(T;\theta \odot m  \right)$
		\Comment{Fine-tune the sparse network with mask}
	\end{algorithmic}
	\label{algorithm1}
\end{algorithm}

\begin{figure*}[htbp]	
	\centering
	\includegraphics[trim={0 0.1cm 0 0.2cm},clip,width=12.87cm]{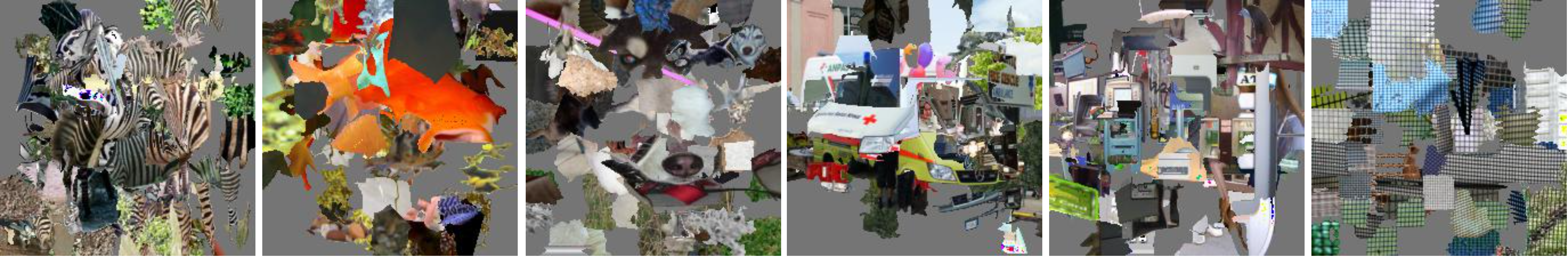}
	
	\caption{Super Stitching with $\sigma=0.75$. From left to right: zebra, goldfish, Siberian husky, ambulance, cash machine, and window screen.}
	\label{stitching}
	%\vspace{-2em}
\end{figure*}

During extracting visual concepts, based on the trained classifier, we derive concept activation vectors (CAVs) \cite{Kim_Wattenberg_Gilmer_Cai_Wexler_Viegas_Sayres_2017,Ghorbani_Wexler_Zou_Kim_2019,Yeh_Kim_Arik_Li_Pfister_Ravikumar_2019} as the normal to a hyperplane that separates concept and random samples.  By computing the similarity between the loss gradient and the CAV, we can finally estimate the importance rating of each concept conditioned on a certain target label. Specially, the importance rating of a certain concept $c$, also called the TCAV (Test with CAVs) score, is defined as follows,
\begin{equation}
	\mathrm{TCAV}_{c}=\frac{\left | \left \{ x_{i} \in X_{k},c\in  x_{i}:\bigtriangledown h^{l}_{k} \left ( F_{l}\left ( c \right ) \right ) \cdot v_{c}^{l}>0\right \} \right |}{\left | X_{k} \right |},
\end{equation}
where $v_{c}^{l}$ is the binary linear classifier of $c$ with random samples at the $l$ layer of $F(T)$. $ F_{l}\left ( c \right )$ is $c$'s activation in $l$, and $\bigtriangledown h^{l}_{k}(F_{l}\left ( c \right ))$ is the logit for $c$ towards class $k$ in $l$.

Importance of each discriminative image patch is evaluated by the TCAV score. As shown in Figure \ref{concepts}, we extract and rank the discriminative image patches. After this, we sample mini-batches in discriminative image patches and then prune the network using a specific OPaI pruning criterion $\mathcal{A}$, which is shown in Algorithm \ref{algorithm1}. We name this pruning algorithm Discriminative One-shot Network Pruning (DOP).

\subsection{Super Stitching}
\label{Super_Stitching}

In addition to DOP, we propose a further data-preprocessing strategy, Super Stitching. This method uses the data to get strengthened gradient information in pruning. Firstly, we define the coverage of valid pixels in each discriminative image patch, $r(c_{k,p,q} )$, as follows,
\begin{equation}
	r\left(c_{k,p,q} \right)=\frac{ \left |  c_{k,p,q}  \right |}{\left | x_{i}  \right |}, \quad c_{k,p,q} \in x_{i},
\end{equation}
which is a ratio of the number of valid pixels in the discriminative image patch over the total number of pixels in the image.
\begin{algorithm}[H]
	\caption{Get Super Stitching Image Patch}
	\begin{algorithmic}[1]
		\Require{$c_{k}$ with $p$ concepts and $q$ segments per concept, threshold $\sigma$}
		
		\State $\mathbb{C}_{k}=0$, $\mathrm{waiting\_list}=\emptyset $
		\State Sort concepts $c_{k,p}$ by importance in ascending order. 
		\While {$r\left(\mathbb{C}_{k}\right) < \sigma$}
		\If{$\mathrm{waiting\_list}==\emptyset $}

		\State $\mathrm{waiting\_list}=c_{k}$
		\Comment{Ensure waiting list is not empty.}
		
		\EndIf
		\For{idx in $ \{1,...,p\}$}
		\State temp $=\mathrm{RandomPop}(c_{k,\mathrm{idx}})$
		\Comment{Random Pop from concept $c_{k,\mathrm{idx}}$.}
		\State $\mathbb{C}_{k}+=$temp
		\Comment{Stitch segment into target image patch.}
		\EndFor
		\EndWhile
		\Return $\mathbb{C}_{k}$	
	\end{algorithmic}
	\label{algorithm2}
\end{algorithm}
Second, we set a hyperparameter threshold $\sigma$ to ensure that the coverage of valid pixels in each stitching patch is larger than $\sigma$. Then, we can obtain the Super Stitching discriminative image patches in the same class from Algorithm \ref{algorithm2}, noted as $\mathbb{C}_{k}$. Since we keep the same position of patches as the original images, when stitching a new patch into $\mathbb{C}_{k}$, overlap may occur. So, we sort all the patches belonging to different concepts in ascending order by their importance. Then, we randomly select a single segment from each concept in turn, and stitch them into $\mathbb{C}_{k}$.  In this way, we make important patches as visible as possible. This process will be repeated until the threshold $\sigma$ is exceeded. In the Figure \ref{stitching} we show several examples of Super Stitching with $\sigma=0.75$. This algorithm ensures that the coverage of all samples obey a Gaussian distribution.
\begin{table}[]
	\small
	\centering
	\begin{tabu}{lcccc}
		\hline
		\multicolumn{1}{l}{Sparsity percentage} & 60\%                              & 80\%                              & 90\%                              & 95\%                              \\
		(Baseline)                              & 76.47\%                           &                                   &                                   &                                   \\ \hline
		SNIP                             & 74\%                              & 70.94\%                           & 61.06\%                           & 36.43\%                           \\
		SNIP with DOP (Ours)                    & \textbf{74.29\%} & \textbf{71.15\%} & \textbf{64.12\%} & \textbf{48.14\%} \\ \hline
		GraSP                           & 73.87\%                           & 71.14\%                           & 67.07\%                           & \textbf{61.76\%}                           \\
		GraSP with DOP (Ours)                   & \textbf{74.19\%}                           & \textbf{71.76\%}                           & \textbf{67.65\%}                           & 60.02\%                          \\ \hline
	\end{tabu}
	\vspace{0.5em}
	\caption{\label{tab:table1}Top-1 Test Accuracy of ResNet-50 on ImageNet.}

\end{table}

Finally, we sample mini-batches from Super Stitching discriminative image patches $\mathbb{C}_{T}^{b}=\{ ( \mathbb{C}_{k},y_{i}  )\}_{i=1}^{b} \sim \mathbb{C}_{T}$ and prune the network.

\section{Experiments}
\label{experiments}
\label{sec:Experiments}
We evaluate DOP and Super Stitching on benchmark ImageNet-1k \cite{Deng_Dong_Socher_Li_Li_Fei-Fei_2009} with ResNet-50 architecture \cite{He_Zhang_Ren_Sun_2015}. Our results provide clear evidence that OPaI is data-dependent, and pruning based on discriminative image patches improves the performance of OPaI.

\subsection{DOP: Pruning ResNet-50 with Varying Levels of Sparsity}
\label{ssec:varying levels of sparsity}

We first compare the performances of DOP  with the original OPaI methods at different sparsity levels. We prune an initialized ResNet-50 network by discriminative image patches. Those discriminative image patches are from a pre-trained ResNet-50 on ImageNet. To implement SNIP and GraSP, we adopt the open codes of Wang \etal \cite{Wang_Zhang_Grosse_2020} and Su \etal \cite{Su_Chen_Cai_Wu_Gao_Wang_Lee_2020}.

Based on the TCAV score, we select the Top-5 important concepts of each class. In the SNIP-based comparison experiments, we select 10,000 materials, i.e., 10 per class. Our sampling strategy follows \cite{Frankle_Dziugaite_Roy_Carbin_2020}, since \cite{Lee_Ajanthan_Torr_2018} do not prune ResNet-50 on ImageNet. In the GraSP-based comparison experiments, we adopt 30,000 materials, i.e., 30 per class, closed to \cite{Wang_Zhang_Grosse_2020}'s implementation (250 batch size with 150 iterations). To sum up, for SNIP with DOP, we randomly select 2 discriminative image patches per concept. For GraSP with DOP, we select 6 discriminative image patches. These discriminative image patches use the same data augmentation as the training set, following \cite{Wang_Zhang_Grosse_2020,Lee_Ajanthan_Torr_2018}. We train the pruned network 120 epochs. The batch size is 128. The optimization is SGD with a learning rate of 0.1 at the beginning, multiplying 0.1 at epochs $\left[30,60,90\right]$, a momentum of 0.9, and a weight decay of 0.0001.

\begin{figure}[!htbp]
	\begin{minipage}[b]{0.45\textwidth}  
		%	\begin{figure}
			\pgfplotsset{ymin=35,ymax=76}
			\centering
			\begin{adjustbox}{max width=\textwidth}
				\begin{tikzpicture}[spy using outlines={circle, magnification=3, size=1.8cm, connect spies}]
					\begin{axis}[
						grid=both,
						xlabel=\textsc{Sparsity percentage (\%)},
						ylabel=\textsc{Top-1 Accuracy (\%)},
						%xmode=log,
						ymode=log,
						log ticks with fixed point,
						%minor xtick={50,...,100},
						%minor ytick={30,...,80},
						legend pos=south west,
						legend style={font=\footnotesize},
						cycle list name=exotic
						]
						\addplot[color=skyblue1,line width=1.2pt,opacity=0.8] plot coordinates {
							(60,     74.29)
							(80,    71.15)
							(90,    64.33)
							(95,   48.14)
						};
						
						\addplot[color=chameleon1,line width=1.2pt,opacity=0.8] plot coordinates {
							(60,      74)
							(80,    70.94)
							(90,    61.06)
							(95,   36.43)
						};
						\addplot[color=butter1,line width=1.2pt,opacity=0.8] plot coordinates {
							(60,     71.75)
							(80,    62.86)
							(90,    48.31)
							(95,   14.62)
						};
						\addplot[color=chocolate1,line width=1.2pt,opacity=0.8] plot coordinates {
							(60,     74.25)
							(80,    71.04)
							(90,    64.25)
							(95,   45.93)
						};
						
						\addplot[color=plum1,line width=1.2pt,opacity=0.8] plot coordinates {
							(60,     74.55)
							(80,    71.58)
							(90,    62.78)
							(95,   41.83)
						};
						\begin{scope}
							\spy[black!70!black,size=2cm] on (5.45,4.3) in node [fill=white] at (1.8,3.5);
						\end{scope}
						
						\legend{SNIP with DOP\\SNIP\\All-One Matrix\\Random Segment\\Less Patch\\}
						
					\end{axis}
				\end{tikzpicture}
			\end{adjustbox}
			%\captionof{\label{fig:ablation2}Ablations at concept maps}
			%\caption{\label{fig:ablation2}Ablations at concept maps}
			%\end{figure}
		\end{minipage}  
		\begin{minipage}[b]{0.45\textwidth}  
			%	\begin{figure}
				\pgfplotsset{ymin=39,ymax=75}
				\centering
				\begin{adjustbox}{max width=\textwidth}
					\begin{tikzpicture}[spy using outlines={circle, magnification=3, size=1.8cm, connect spies}]
						\begin{axis}[
							grid=both,
							xlabel=\textsc{Sparsity percentage (\%)},
							ylabel=\textsc{Top-1 Accuracy (\%)},
							%xmode=log,
							ymode=log,
							log ticks with fixed point,
							%minor xtick={50,...,100},
							%minor ytick={30,...,80},
							legend pos=south west,
							legend style={font=\footnotesize},
							cycle list name=exotic
							]
							\addplot[color=skyblue1,line width=1.2pt,opacity=0.8] plot coordinates {
								(60,     73.99)
								(80,    71.53)
								(90,    66.96)
								(95,   56.92)
							};
							
							\addplot[color=chameleon1,line width=1.2pt,opacity=0.8] plot coordinates {
								(60,      73.57)
								(80,    71.01)
								(90,    67.37)
								(95,   62.29)
							};
							\addplot[color=butter1,line width=1.2pt,opacity=0.8] plot coordinates {
								(60,     73.57)
								(80,    63.85)
								(90,    42.77)
								(95,   11.06)
							};
							\addplot[color=chocolate1,line width=1.2pt,opacity=0.8] plot coordinates {
								(60,     73.92)
								(80,    71.37)
								(90,    67.06)
								(95,   53.6)
							};
							
							\addplot[color=plum1,line width=1.2pt,opacity=0.8] plot coordinates {
								(60,     73.9)
								(80,    50.66)
								(90,    39.82)
								(95,   30.87)
							};

							\begin{scope}
								\spy[black!70!black,size=2cm] on (5.45,4.6) in node [fill=white] at (1.8,3.5);
							\end{scope}
							
							\legend{GraSP with DOP\\GraSP\\All-One Matrix\\Random Segment\\Less Patch\\}
							
						\end{axis}
					\end{tikzpicture}
				\end{adjustbox}
				%\captionof{\label{fig:ablation2}Ablations at concept maps}
				%\caption{\label{fig:ablation2}Ablations at concept maps}
				%\end{figure}
				%\captionof{figure}{Some figure}
				%c
				
			\end{minipage}  
			\caption{Ablation experiments at DOP. \textbf{Left:} SNIP with DOP. \textbf{Right:} GraSP with DOP. If the OPaI is data-independent, then the same results should be obtained in the ablation experiments. However, we still observe disparity in ablation experiments. Such trends indict that OPaI is impacted by limited validation information. }
			\label{fig:ablation}

		\end{figure}
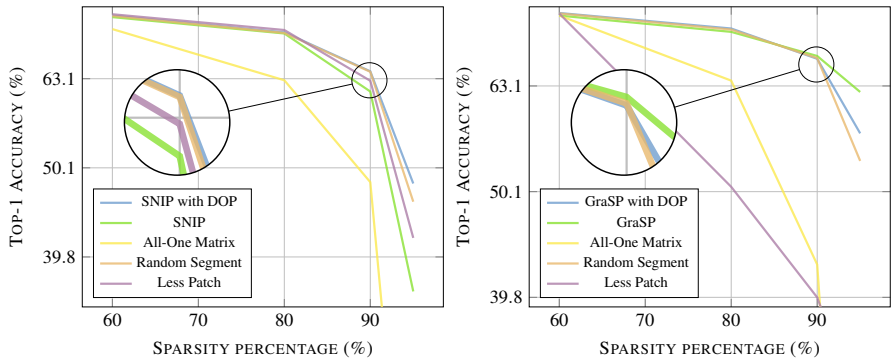
	
The results are shown in the Table \ref{tab:table1}. Sparsity percentage is the proportion of parameters which are equal to 0 in the overall model parameters. Based on SNIP pruning criterion, we show that our DOP dominates the original SNIP at all four sparsity levels \{60\%, 80\%, 90\%, 95\%\} at Top-1 accuracy performances. When the sparsity reaches 90\%, the gap has widened to about 3\%. Moreover, based on GraSP pruning criterion, our DOP has a slight advantage at 90\% sparsity and fails at 95\% sparsity. 

The results indicate that, for SNIP and GraSP, the discriminative image patches benefit the pruning performance. The failure of GraSP with DOP at 95\% sparsity could be owing to a lack of valid information. We verify the guess in ablation experiments in Section \ref{ssec:Ablations} and propose Super Stitching in Section \ref{superstitchingexp} to solve the problem.

		\subsection{Ablation Experiments on Discriminative Image Patches}
		\label{ssec:Ablations}
		We design three ablation experiments to investigate the function of data in OPaI by gradually changing the content of input data. If the same experimental results are obtained with various input data, then the data-independence can be verified; otherwise OPaI is data-dependent.   The three ablation experiments are designed as follows.
		
		\textbf{\textit{All-One Matrix}} explores the case where the image has no content in pruning. In this experiment, we prune the network using all-one matrices with the same size and labels as the original images. If \textit{All-One Matrix} performs similarly to SNIP and GraSP,  it proves the data-independence in OPaI; otherwise, OPaI should be data-dependent. This attempt is similar to the work of Tanaka \etal \cite{Tanaka_Kunin_Yamins_Ganguli_2020}, who devise a fully data-independent iterative pruning method with a different pruning criterion to avoid layer collapse. However, we explore the impact of image content on OPaI, which is pretty different from the purpose of \cite{Tanaka_Kunin_Yamins_Ganguli_2020}.

		\textbf{\textit{Random Segment}} considers the segmentation impact on OPaI. We adopt the SLIC algorithm \cite{Achanta_Shaji_Smith_Lucchi_Fua_Su} used by ACE to segment the images, and randomly select the same number of segments. We aim to confirm that the original images could include not only discriminative contents, but also image segments that are meaningless for pruning.

		\textbf{\textit{Less Patch}} explores the influence of the material number on pruning  when informative data become less. Only one discriminative image patch is selected for each class in the pruning step, i.e., a total of 1,000 discriminative image patches.

		Except \textit{Less Patch}, all ablation experiments adopt the same number of materials to guarantee a fair comparison, i.e., 10,000 materials.  The results are shown in Figure \ref{fig:ablation}. \textit{All-One Matrix} performs worse in all trails. As the sparsity increases, the gap between  \textit{All-One Matrix} and the network pruned by the image or discriminative image patches becomes wider.  \textit{Random Segment} performs similarly to DOP at lower sparsity but worse than DOP at higher sparsity. It verifies our guess that the images include both useful discriminative information and non-useful information for pruning.  In \textit{Less Patch}, the performance is worse than DOP, and GraSP is more sensitive to the amount of informative data than SNIP. GraSP with  \textit{Less Patch} is even worse than  \textit{All-One Matrix} at some sparsity levels. It confirms that our experimental results in Section \ref{ssec:varying levels of sparsity}, namely, GraSP requires more helpful information at high sparsity to prune.
		
			\begin{table}[]
			\small
			\centering
			\resizebox{\textwidth}{27mm}{
				\begin{tabular}{l|ll|llll}
					\hline
					\multirow{2}{*}{Method}           & \multirow{2}{*}{Material Number} & \multirow{2}{*}{Material Type}                                                       & \multicolumn{4}{c}{Sparsity percentage}                                   \\ \cline{4-7} 
					&                                  &                                                                                      & 60\%             & 80\%             & 90\%             & 95\%             \\ \hline
					GraSP                             & 30,000                           & image                                                                                & 73.87\%          & 71.14\%          & 67.07\%          & 61.76\%          \\
					GraSP with DOP (Ours)             & 30,000                           & discriminative patch                                                                 & \textbf{74.19\%} & \textbf{71.76\%} & 67.65\%          & 60.02\%          \\
					GraSP with Super Stitching (Ours) & 30,000                           & \begin{tabular}[c]{@{}l@{}}stitching patch\\ $\sigma=0.5$\end{tabular}  & 74.14\%          & 71.57\%          & 67.59\%          & \textbf{62.70\%} \\
					GraSP with Super Stitching (Ours) & 10,000                           & \begin{tabular}[c]{@{}l@{}}stitching patch\\ $\sigma=0.75$\end{tabular} & 73.94\%          & 71.65\%          & \textbf{68.02\%} & 62.06\%          \\ \hline
					GraSP (Wang \etal \cite{Wang_Zhang_Grosse_2020})                      & 37,500                           & image                                                                                & 74.02\%          & 72.06\%          & 68.14\%          & -                \\
					GraSP (Jorge \etal \cite{Jorge_Sanyal_Behl_Torr_Rogez_Dokania_2020})                     & 614,440                          & image                                                                                & -                & -                & 65.4\%           & 46.2\%           \\
					GraSP (Frankle \etal \cite{Frankle_Dziugaite_Roy_Carbin_2020})                   & 10,000                           & image                                                                                & 73.4\%           & 71.0\%           & 67\%             & -                \\
					GraSP (Hayou \etal \cite{Hayou_Ton_Doucet_Teh_2020})                     & -                                & image                                                                                & -                & -                & 66.41\%          & 62.1\%           \\
					FORCE (Jorge \etal \cite{Jorge_Sanyal_Behl_Torr_Rogez_Dokania_2020})                     & 614,440                          & image                                                                                & -                & -                & 64.9\%           & 59.0\%           \\
					Iter SNIP (Jorge \etal \cite{Jorge_Sanyal_Behl_Torr_Rogez_Dokania_2020})                 & 614,440                          & image                                                                                & -                & -                & 63.7\%           & 54.7\%           \\
					SynFlow (Hayou \etal \cite{Hayou_Ton_Doucet_Teh_2020})                   & -                                & all-one matrix                                                                                & -                & -                & 66.2\%           & 62.05\%          \\
					SBP-SR (Hayou \etal \cite{Hayou_Ton_Doucet_Teh_2020})                    & -                                & image                                                                                & -                & -                & 67.02\%          & 62.66\%          \\
					ProsPr (Alizadeh \etal \cite{Alizadeh_Tailor_Zintgraf_Amersfoort_Farquhar_Lane_Gal_2022})                 & -                                & image                                                                                & -                & -                & 66.86\%          & 59.62\%          \\ \hline
			\end{tabular}}
			\caption{DOP and Super Stitching test accuracy of ResNet-50 on ImageNet based on GraSP pruning criterion. The empty means that the description or experiment is missing. At high sparsity, we can observe that Super Stitching has a significant advantage. Also, DOP has advantages at low sparsity.}
			\label{stitching_res}
		\end{table}
		
		By gradually changing the data input in the OPaI algorithm, we explicitly show that valid information in the OPaI method directly impacts the pruning performance in the experiments. 
		\subsection{Super Stitching Compared to SOTA}
		\label{superstitchingexp}
		We create Super Stitching to further improve the discriminative image patches in OPaI. We aim to use fewer but more informative samples to enhance the gradient flow. In implementation details, we set a coverage threshold $\sigma \in (0.5,0.75)$ to balance the overlap and performance. Figure \ref{stitching} shows the stitched discriminative image patches with $\sigma = 0.75$. We adopt the GraSP pruning criterion, since GraSP achieves better performance for ResNet-50 on ImageNet than SNIP \cite{Frankle_Dziugaite_Roy_Carbin_2020,Wang_Zhang_Grosse_2020,Jorge_Sanyal_Behl_Torr_Rogez_Dokania_2020}. 
		
		The results of Super Stitching experiments are in Table \ref{stitching_res}. The first part of the table shows our pruning experiment results on ImageNet for ResNet-50 with a fixed random seed. At the second part of the table, we report the results in corresponding papers of ResNet-50 on ImageNet, where FORCE \cite{Jorge_Sanyal_Behl_Torr_Rogez_Dokania_2020}, Iter SNIP \cite{Jorge_Sanyal_Behl_Torr_Rogez_Dokania_2020} and SynFlow \cite{Tanaka_Kunin_Yamins_Ganguli_2020} are iterative pruning algorithms. We notice that SBP-SR \cite{Hayou_Ton_Doucet_Teh_2020} claims a higher performance for ResNet-50 on ImageNet recently, which is the SOTA OPaI method. When the sparsity is high, we can observe that Super Stitching has a distinct advantage with fewer samples. Furthermore, DOP has an advantage in low sparsity. It is noticed that DOP and Super Stitching achieved their best results at 60\% and 95\% sparsity, respectively, outperforming the original SOTA. The results demonstrate that discriminative data in OPaI is able to perform better than iterative pruning with less costs.

		\section{Conclusion}
		\label{sec:Conclusion}
		To explore whether data matters in OPaI, we introduce DOP and Super Stitching, two novel data-dependent methods for OPaI based on discriminative image patches. Our research not only reveals that informative data is helpful in OPaI, but it also shows that DOP and Super Stitching can significantly improve pruning performance. This conclusion refreshes our typical views of the OPaI method and provides us with a new route for OPaI advancement. It is especially worth noting that our methods and experiments can help us understand which data are critical for networks at an earlier stage – providing valuable suggestions for PaI improvement.

\bibliography{egbib}
\end{document}